\documentclass[12pt]{article}
\usepackage[margin=2cm]{geometry}
\usepackage[title]{appendix}
\usepackage{tikz, pgfplots}\usetikzlibrary{fadings}
\usetikzlibrary{shapes.multipart,  arrows.meta}
\usepackage{subcaption}

\newtheorem{definition}{Definition}
\newtheorem{remark}{Remark}

\newtheorem{theorem}{Theorem}
\newtheorem{proof}{Proof}

\usepackage{amsmath, amssymb,mathtools}
\usepackage[utf8]{inputenc} 
\usepackage[T1]{fontenc}    
\usepackage{hyperref}       
\usepackage{url}            
\usepackage{booktabs}       
\usepackage{amsfonts}       
\usepackage{nicefrac}       
\usepackage{microtype}      
\usepackage{cleveref}       
\usepackage{lipsum}         
\usepackage{graphicx}
\usepackage{doi}

\title{Backpropagation and F-adjoint}

\author{{Ahmed Boughammoura}${}^*$}

\usepackage{stmaryrd}
\usepackage{nccmath}
\makeatletter
 \newcommand{\xmapsfrom}[2][]{%
 \ext@arrow3095\leftarrowfill@{#1}{#2}\mapsfromchar
}
\makeatother
\usetikzlibrary{decorations.pathreplacing}
    \usetikzlibrary{positioning}
     \usetikzlibrary{calc}
\providecommand{\keywords}[1]{\textbf{\textit{Key words--}} #1}
\begin{document}
\maketitle
\makeatletter{\renewcommand*{\@makefnmark}{}
\footnotetext{${}^*$ Department of Mathematics\\ \hspace*{0.75cm} Higher Institute of Informatics and Mathematics, Monastir, Tunisia\\ \hspace*{0.75cm} {Email: {ahmed.boughammoura@gmail.com}}\makeatother}
\begin{abstract}{This paper presents a concise mathematical framework for investigating both feed-forward and backward process, during the training to learn model weights,  of an artificial neural network (ANN). Inspired from the idea of the two-step rule for backpropagation,  we  define  a notion of F-adjoint which is aimed at a better   description of the backpropagation algorithm. In particular, by introducing the notions of F-propagation and F-adjoint through a deep neural network architecture, the backpropagation associated to a cost/loss function is   proven to be completely characterized by the F-adjoint of the corresponding F-propagation relatively to the partial derivative, with respect to the inputs,  of the cost function.}
\end{abstract}


\section{Introduction}\label{A}

~~~ An Artificial Neural Network (ANN) is a mathematical model which is intended to be a universal function approximator which learns from data (cf. McCulloch and Pitts, \cite{McCulloch1943}). 
In general, an ANN consists of a number of units called artificial neurons, which are  a composition of affine mappings, and non-linear (activation) mappings (applied element wise), connected by weighted connections and organized into layers, containing an input layer, one or more hidden layers, and an output layer.The neurons in an ANN can be connected in many different ways. In the simplest cases, the outputs from one layer are the inputs for the neurons in the next
layer. An ANN  is said to be a feedforward ANN, if outputs from one layer of neurons
are the only inputs to the neurons in the following layer. In a fully connected ANN, all neurons in one layer are connected to all neurons in the previous layer (cf. page 24 of \cite{Baldi2021}). An example of a fully connected feedforward network is presented in Figure \ref{fig:multilayer-perceptron}.

In the present work we focus essentially on feed-forward artificial neural networks, with $L$ hidden layers  and a transfer (or activation) function $\sigma$, and the corresponding  supervised learning problem. Let us define a simple artificial neural network as follows:

\begin{equation}
\label{feedforward-formulas-scalar1}
X^{0}= x,\ 
Y^{h}=W^{h}.X^{h-1},\  
  X^{h}=\sigma(Y^{h}), \ h=1,\cdots, L
\end{equation}
where $ x\in\mathbb{R}^n$ is the input to the network, $h$ indexes the hidden layer and $W^h$ is the weight matrix of the $h$-th hidden layer. In what follows we shall refer to the two equations of \eqref{feedforward-formulas-scalar1} as the two-step recursive forward formula. The two-step recursive forward formula is very useful in obtaining the outputs of the feed-forward deep neural networks.

A major empirical issue in the neural networks is to estimate the unknown parameters
$W^h$ with a sample of data values of targets and inputs. This estimation procedure is characterized by the recursive updating or the learning of estimated parameters. This algorithm is called the   backpropagation algorithm. 
 As reviewed by Schmidhuber \cite{Schmidhuber15}, back-propagation was introduced and developed  during the 1970’s and 1980’s and refined  by Rumelhart et al. \cite{rumelhart1986}). In addition, it is well known that  the most important algorithms of artificial neural networks training is the back-propagation algorithm. From mathematical point view,  back-propagation is a method  to minimize errors for a loss/cost function through gradient descent.  More precisely, an input data is fed to the network and forwarded through the so-called layers ; the produced output is then fed to the cost function to compute the    gradient of the associated error. The computed gradient is then back-propagated through the layers to update the weights by using the well known gradient descent algorithm.

As explained in \cite{rumelhart1986}, the goal of back-propagation is to compute the partial derivatives  of the cost function $J$. In this procedure,  each hidden layer $h$ is assigned teh so-called delta error term $\delta^h$. For each hidden layer, the delta error term $\delta^h$ is derived from the delta error terms   $\delta^k,\ k=h+1,\cdots L$; thus the concept of error back-propagation. The output layer $L$ is the only layer
whose error term $\delta^L$ has no error dependencies, hence $\delta^L$  is then given by the following equation
\begin{equation}
\label{delta-L}
\delta^L=\frac{\partial J}{\partial W^L}\odot \sigma'(Y^{L}),
\end{equation}
where $\odot$  denotes the element-wise matrix multiplication (the so-called Hadamard product, which is exactly the element-wise multiplication  $"*"$ in Python).
For the error term $\delta^h$, this term is derived from matrix multiplying $\delta^{h+1}$  with the weight transpose matrix $(W^{h+1})\top$ and subsequently
multiplying (element-wise) the activation function derivative $\sigma'$ with respect to the preactivation $Y^{h}$. Thus, one has the following equation
\begin{equation}
\label{delta-l}
\delta^h=(W^{h+1})^\top\delta^{h+1}\odot \sigma'(Y^{h}),\ h=(L-1),\cdots, 1
\end{equation}
Once the layer error terms have been assigned, the partial derivative $\frac{\partial J}{\partial W^h}$ can be computed by
\begin{equation}
\label{delta-W}
\frac{\partial J}{\partial W^h}=\delta^{h+1}(X^{h})^\top
\end{equation}
In particular, we deduce that the back-propagation algorithm is uniquely  responsible for computing weight partial
derivatives of $J$ by using the recursive equations \eqref{delta-W} and \eqref{delta-l} with the initialization data given by \eqref{delta-L}. This procedure is often called "the generalized delta rule".
The key question to which we address ourselves in the present work is the following:  how could one reformulate  this "generalized delta rule" in two-step recursive backward formula as  \eqref{feedforward-formulas-scalar1} ?
 
~~~ In the present work, we shall provided a concise mathematical answer to the above question. In particular, we shall introduce the so-called two-step rule for back-propagation, recently
proposed by the author in \cite{bougham2023},  similar to the one for forward propagation. Moreover, we explore some mathematical concepts behind the two-step   rule for backpropagation. 
 

The rest of the paper is organized as follows.  
Section 2 outlines some notations, setting and ANN framework. In Section 3 we recall and develop the two-step rule for back-propagation. In Section 4 we introduce the   concepts of F-propagation and the associated F-adjoint and rewrite the two-step rule with these notions.
In  Section 5  we provide some application of this method to study some simple cases. In Section 6 conclusion, related works and mention future work directions are given.

\section{Notations, Setting  and the ANN} 
~~~Let us now precise some notations. Firstly, we shall denote  any vector $X \in \mathbb{R}^n$, is considered as columns $X=\left(X_1,\cdots, X_{n} \right)^\top$ and for any  family of transfer functions $\sigma_i: \mathbb{R}\rightarrow \mathbb{R},\ i=1,\cdots, n$, we shall introduce the coordinate-wise map $\sigma : \mathbb{R}^n \rightarrow \mathbb{R}^n$   by the following formula
\begin{equation}
\label{transfer-formulas}
\sigma(X):=\left(\sigma_1(X_1),\cdots, \sigma_n(X_{n}) \right)^\top.
\end{equation}
This map   can be considered as an
“operator” Hadamard multiplication of columns $\sigma=\left(\sigma_1,\cdots, \sigma_{n} \right)^\top$ and 
$X=\left(X_1,\cdots, X_{n} \right)^\top$, i.e., 
$\sigma(X)=\sigma\odot X.$
Secondly, we shall need to recall some useful multi-variable
functions derivatives notations. For any $n,m\in \mathbb{N}^*$ and any differentiable function  with respect to the variable $x$
\begin{equation}
F:\ \mathbb{R}\ni x\mapsto F(x)=\Bigl(F_{ij}(x)\Bigr)_{\substack{1\leq i\leq m \\ 1\leq j\leq n }}  \in \mathbb{R}^{m\times n}
\end{equation} we use the following notations associated to the partial derivatives of  $F$ with respect to $x$ 
\begin{equation}
	 \frac{\partial F}{\partial x} =
		\left(
		\frac{\partial F_{ij}(x)}{\partial x} 
	\right)_{\substack{1\leq i\leq m \\ 1\leq j\leq n }} 
	\end{equation} 
	
In adfdition, for any $n,m\in \mathbb{N}^*$ and any differentiable function  with respect to the matrix variable
\begin{equation}
F:\ \mathbb{R}^{m\times n}\ni X=\Bigl(X_{ij}\Bigr)_{\substack{1\leq i\leq m \\ 1\leq j\leq n }}\mapsto F(X) \in \mathbb{R}
\end{equation}

 we shall use the so-called  \textbf{denominator layout} notation (see page 15 of \cite{Ye2022}) for the partial derivative of  $F$ with respect to the matrix  $X$ 
\begin{equation}
	 \frac{\partial F}{\partial X} =
		\left(
		\frac{\partial F(X)}{\partial X_{ij}} 
	\right)_{\substack{1\leq i\leq m \\ 1\leq j\leq n }} 
	\end{equation}

In particular, this notation leads to the following useful formulas: for any $q\in\mathbb{N}^*$ and any matrix $W\in \mathbb{R}^{q\times m}$ we have
 \begin{equation}\label{usefl-x}
	 \frac{\partial (WX)}{\partial X} =W^\top, 
	\end{equation}
	when $X\in \mathbb{R}^{ n}$ with $X_n=1$ one has
 \begin{equation}\label{usefl-w-sharpe}
	 \frac{\partial (WX)}{\partial X} =W_\sharp^\top
	\end{equation} 
	where $W_\sharp$ is the matrix $W$ whose last column is removed (this formula  is highly useful in practice). 
Moreover,	for any   matrix $X\in \mathbb{R}^{m\times n}$ we have
 \begin{equation}\label{usefl-w}
	 \frac{\partial (WX)}{\partial W} =X^\top.
	\end{equation}
	
Then, by the chain rule one has
for any $q,n,m\in \mathbb{N}^*$ and any differentiable function  with respect to the matrices variables $W,X$ :
\begin{equation*}
F:\ (W, X)\mapsto Z:=WX\in \mathbb{R}^{q\times n}\mapsto F(Z) \in \mathbb{R}
\end{equation*}
	\begin{equation}\label{chain:}
	 \frac{\partial F}{\partial X} =
				W^\top\frac{\partial F}{\partial Z} \ 
				\mathrm{and}\ 
				\frac{\partial F}{\partial W} =
				\frac{\partial F}{\partial Z}X^\top.
	\end{equation}
Furthermore, for any   differentiable function  with respect to  $Y$
\begin{equation*}
F:\ \mathbb{R}^n\ni Y\mapsto X:=\sigma(Y)\in \mathbb{R}^{n}\mapsto F(X) \in \mathbb{R}
\end{equation*} we have
	\begin{equation}\label{chain:2}
	 \frac{\partial F}{\partial Y} =
				 \frac{\partial F}{\partial X}\odot\sigma'(Y).
	\end{equation}	
Throughout this paper, we consider layered feedforward neural networks and supervised learning tasks. Following \cite{Baldi2021} (see (2.18) in page 24), we will denote such an architecture by
\begin{equation}
\label{architectue}
A[N_0, \cdots, N_h,\cdots, N_L]
\end{equation}
where $N_0$ is the size of the input layer, $N_h$ is the size of hidden layer $h$,
and $N_L$ is the size of the output layer; $L$ is defined as the depth of the ANN, then the neural network is called as Deep Neural Network (DNN).
 We assume that the layers are fully
connected, i.e.,  neurons between two adjacent layers are fully pairwise connected, but neurons within a single layer share no connections. 

In the next section we shall recall, develop and improve the two-scale rule for backpropagation, recently introduced by the author  in \cite{bougham2023}.
\section{Two-step rule for backpropagation}

First, let us mention that the two-step rule for backpropagation 
 is very useful in obtaining  some estimation of the unknown parameters $W_h$ with a sample of data values of targets and inputs (see some examples given in \cite{bougham2023}).   
 
Now, let $\alpha_{ij}^h$ denote the weight connecting neuron $j$ in layer $h-1$ to neuron $i$ in hidden layer $h$ and let the associated transfer function denoted $\sigma_i^h$. In general, in the application two different passes of computation are distinguished. The first pass is referred to as the forward pass, and the second is referred to as the backward pass. In the forward pass, the synaptic weights remain fixed throughout the network, and  the output  $X_i^h$ of neuron $i$ in hidden layer $h$ is computed by the following recursive-coordinate form :
\begin{equation}
\label{feedforward-formulas-scalar}
X_i^h:=\sigma_i^h(Y_i^{h})\quad\mathrm{where}\quad Y_i^{h}:=\sum_{j=1}^{N_{h-1}}\alpha_{ij}^h X_j^{h-1}
\end{equation}

In two-step recursive-matrix form, one may rewrite the above formulas as
\begin{equation}
\label{feedforward-formulas-matrix}
X^h=\sigma^h(Y^{h})\quad\mathrm{where}\quad Y^{h}=W^hX^{h-1},
\end{equation} with
\begin{equation}
\label{feedforward-formulas-matrix_with}
 \sigma^h:=(\sigma_1^h,\cdots,\sigma_{N_h}^h)^T,\  W^h:=(\alpha_{ij}^h)\in\mathbb{R}^{N_h}\times\mathbb{R}^{N_{h-1}}.\end{equation}

 Let us assume that for all $h\in\{ 1,\cdots, L\}$,  $\sigma^h=(\sigma,\cdots,\sigma)^\top$, where $\sigma$ is a fixed activation function, and define a simple artificial neural network as follows:

\begin{equation}
\label{feedforward}
  {\mathrm{given}}\ X^{0}=x,\ {\mathrm{for}}\ h=1,\cdots, L,\ 
Y^{h}=W^{h}.X^{h-1},\  X^{h}=\sigma(Y^{h}).
\end{equation}

where $ x\in\mathbb{R}^n$ is the input to the network. 

To optimize the neural network, we compute the partial derivatives of the cost $J(.)$ w.r.t. the weight matrices $\frac{\partial J(.)}{\partial W^h}$. This quantity can be computed by making use of the chain rule in the back-propagation algorithm. To compute the partial derivative with respect to the matrices variables $\{{X^h},{Y^h}, {W^h}\}$, we put 
\begin{equation}
\label{notations}
{X^h_*} =\frac{\partial J(.)}{\partial {X^h}},\ 
{Y^h_*}  =\frac{\partial J(.)}{\partial {Y^h}},\
\delta_{W^{h}}  =\frac{\partial J(.)}{\partial {W^h}}.
\end{equation}

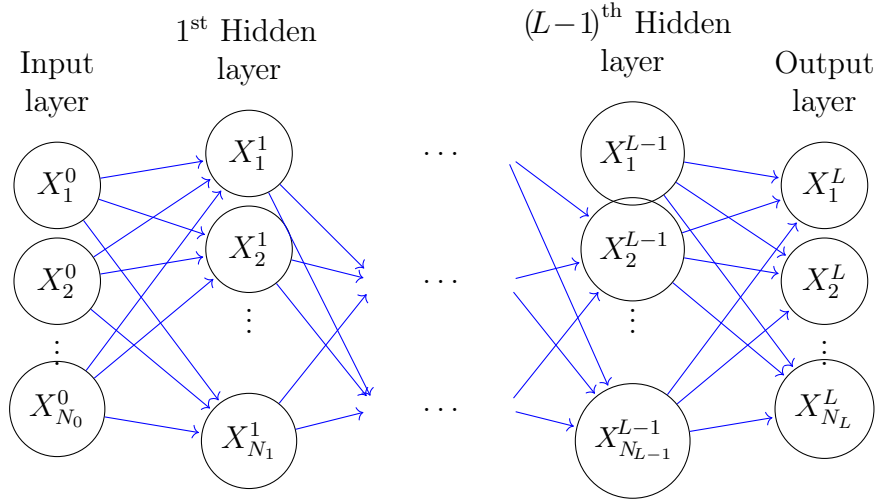
\begin{figure}[htp]
	\centering
	\begin{tikzpicture}[shorten >=1pt,scale=0.85]]
		\tikzstyle{unit}=[draw,shape=circle,minimum size=1.15cm]
		\tikzstyle{hidden}=[draw,shape=circle,minimum size=1.15cm]
		\node[unit](x0) at (0,3.5){$X_1^0$};
		\node[unit](x1) at (0,2){$X_2^0$};
		\node at (0,1){\vdots};
		\node[unit](xd) at (0,0){$X_{N_0}^0$};
		\node[hidden](h10) at (3,4){$X_1^{1}$};
		\node[hidden](h11) at (3,2.5){$X_2^{1}$};
		\node at (3,1.5){\vdots};
		\node[hidden](h1m) at (3,-0.5){$X_{N_{1}}^{1}$};
		\node(h22) at (5,0){};
		\node(h21) at (5,2){};
		\node(h20) at (5,4){};
		\node(d3) at (6,0){$\ldots$};
		\node(d2) at (6,2){$\ldots$};
		\node(d1) at (6,4){$\ldots$};
		\node(hL12) at (7,0){};
		\node(hL11) at (7,2){};
		\node(hL10) at (7,4){};
		\node[hidden](hL0) at (9,4){$X_1^{\!L-1\!}$};
		\node[hidden](hL1) at (9,2.5){$X_2^{\!L-1\!}$};
		\node at (9,1.5){\vdots};
		\node[hidden](hLm) at (9,-0.5){\small{$X_{N_{\!L-1\!}}^{\!L-1\!}$}};
		\node[unit](y1) at (12,3.5){$X_1^{L}$};
		\node[unit](y2) at (12,2){$X_2^{L}$};
		\node at (12,1){\vdots};	
		\node[unit](yc) at (12,0){$X_{N_L}^{L}$};
		\draw[->, blue] (x0) -- (h10);
\draw[->, blue] (x0) -- (h11);
		\draw[->, blue] (x0) -- (h1m);
\draw[->, blue] (x1) -- (h10);
		\draw[->, blue] (x1) -- (h11);
		\draw[->, blue] (x1) -- (h1m);
		\draw[->, blue] (xd) -- (h11);
\draw[->, blue] (xd) -- (h10);
		\draw[->, blue] (xd) -- (h1m);
		\draw[->, blue] (hL0) -- (y1);
		\draw[->, blue] (hL0) -- (yc);
		\draw[->, blue] (hL0) -- (y2);
		\draw[->, blue] (hL1) -- (y1);
		\draw[->, blue] (hL1) -- (yc);
		\draw[->, blue] (hL1) -- (y2);
		\draw[->, blue] (hLm) -- (y1);
		\draw[->, blue] (hLm) -- (y2);
		\draw[->, blue] (hLm) -- (yc);
		\draw[->,path fading=east, blue] (h10) -- (h21);
		\draw[->,path fading=east, blue] (h10) -- (h22);	
		\draw[->,path fading=east, blue] (h11) -- (h21);
		\draw[->,path fading=east, blue] (h11) -- (h22);
		\draw[->,path fading=east, blue] (h1m) -- (h21);
		\draw[->,path fading=east, blue] (h1m) -- (h22);
		\draw[->,path fading=west, blue] (hL10) -- (hL1);
		\draw[->,path fading=west, blue] (hL11) -- (hL1);
		\draw[->,path fading=west, blue] (hL12) -- (hL1);
		\draw[->,path fading=west, blue] (hL10) -- (hLm);
		\draw[->,path fading=west, blue] (hL11) -- (hLm);
		\draw[->,path fading=west, blue] (hL12) -- (hLm);
\node[above=5pt of x0, align=center](x0) {Input\\ layer};
		\node[above=5pt of h10, align=center] (h10) {${\!1}^{\text{st}}$ Hidden \\ layer};
		\node[above=5pt of hL0, align=center] (hL0){${\!\!(\!L\!-\!1\!)}^{\text{th}}$ Hidden \\ layer};
		\node[above=5pt of y1, align=center] (y1){Output\\ layer};
	\end{tikzpicture}
	\caption[Example of a $(L+1)$-layer perceptron.]{Example of an $A[N_0,\cdots, N_L]$ architechture.}
	\label{fig:multilayer-perceptron}
\end{figure}
Let us notice that the forward pass computation between the two adjacent layers $h-1$ and $h$ may be represented mathematically as the composition of the following two maps:

\begin{align*}
{}\hskip1em  \mathbb{R}^{N_{h-1}} 
&\xrightarrow[\hskip4em]{}
\mathbb{R}^{N_{h}}\ \xrightarrow[\hskip4em]{}
\mathbb{R}^{N_{h}}
 \\
X^{h-1}
&\xmapsto[\hskip4em]{W^h}  
  Y^h\ \xmapsto[\hskip5em]{\sigma}  
 X^h
\end{align*}
The backward   computation between the two adjacent layers $h$ and $h-1$ may be represented mathematically as follows:

\begin{align*}
{}\qquad\qquad\mathbb{R}^{N_{h-1}} 
&\xleftarrow[\hskip4em]{}
 %
\mathbb{R}^{N_{h}}
\xleftarrow[\hskip4em]{}
\mathbb{R}^{N_{h}}\qquad
 \\
 X^{h-1}_* 
&\xmapsfrom[\hskip4em]{(W^{h})^\top{(.)}}   
{Y^{h}_*}
\xmapsfrom[{\hskip4em}]{(\odot){\sigma}'(Y^h)}   
{X^{h}_*}
\end{align*}

One may represent those maps  as a simple  mapping diagrams with $X^{h-1}$ and   $X^{h}_*$ as inputs and  the respective outputs are    $Y^h={W^h}X^{h-1},X^h=\sigma(Y^h)$  and $Y^{h}_*, X^{h-1}_*$ (see Figure \ref{fig:forward-pass}).
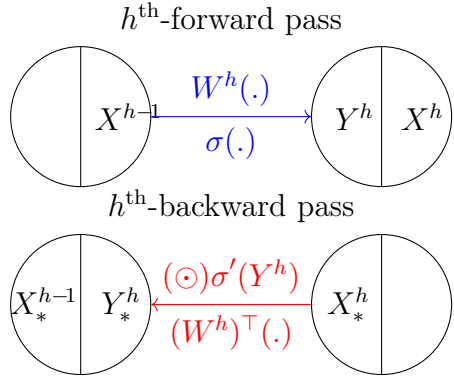
\begin{figure}[htp]
	\centering
\begin{tikzpicture}[xscale=2.5,yscale=1.0, node distance=5cm]
    \tikzstyle{neuron}=[circle,draw=black,minimum size=30pt,inner sep=0pt]
    \tikzstyle{neuron2}=[%
                       circle split,
                       text width=1.5cm,
                       draw=black,
                       rotate=90
                       ]
\begin{scope}
    \node[neuron2] (I1) at (0,4) {
       \nodepart[text width=1.5cm]{one}
       \nodepart[text width=1cm]{two} } ;     
     \node at (I1) {${}$\hspace{7.ex}$X^{h-\!1}$} ;   
     \node[neuron2] (H1) at (1.6,4) {
       } ;     
     \node at (H1) {$Y^h$\hspace{2ex}$X^h\!\!\!$} ;    
    \draw[->, blue] (I1) -- (H1) node[midway,above](w) {$W^h(.)$}  node[midway,below] {$\sigma(.)$} ;                   
\node[above = 2mm of w] (textforward) {$h^{\mathrm{th}}$-forward pass};
%
\end{scope}
\begin{scope}
    \node[neuron2] (I12) at (0,1.5) {
       \nodepart[text width=1cm]{one}
       \nodepart[text width=1cm]{two} } ;     
     \node at (I12) {\!\!$X^{h-\!1}_*$\hspace{1.ex}$\ Y^{h}_*$} ;   
     \node[neuron2] (H12) at (1.6,1.5) {
       } ;     
     \node at (H12) {${}$\hspace{2ex}$X^{h}_*\quad\qquad$} ;    
    \draw[<-, red] (I12) -- (H12) node[midway,above](s) {${(\odot)\sigma}'(Y^h)$}  node[midway,below] {$(W^h)^\top{(.)}$} ;                   
\node[above   = 2mm of s] {$h^{\mathrm{th}}$-backward pass};
\end{scope}                 
\end{tikzpicture}
\caption[ ]{Mapping diagrams associated to the forward and backward passes between two adjacent layers.}
	\label{fig:forward-pass}
\end{figure}

\begin{remark}\label{rem1}
{}\noindent

It is crucial to remark that, if we impose the following setting on $X^h, W^h$ and $\sigma_{N_h}^h$:
\begin{enumerate}
\item  all input vectors have the form $X^h = [X^h_1 , \cdots , X^h_{{N_h}-1}, 1]^\top$ for all $0\leq h\leq (L-1)$;
\item the last functions $\sigma_{N_h}^h$  in the columns $\sigma^h$ for all $1\leq h \leq (L-1)$ are   constant functions equal to $1$.
\end{enumerate}
Then, the $A[N_0,\cdots, N_L]$ neural network will be equivalent to a $(L-1)$-layered affine
neural network with $(N_0-1)$-dimensional input and $N_L$-dimensional output. Each hidden layer $h$ will contain $(N_h-1)$ “genuine” neurons and one (last) “formal”, associated to the bias; the last column of the matrix $W^h$ will be the bias vector for the $h$-th layer (For more details, see the examples given in Section 5). 
\end{remark}
 Now, we state our first main mathematical result to answer  the above question,  in the following theorem.
 
\begin{theorem}[The two-step rule for backpropagation]
{}\noindent\label{thm1}

Let $L$ be the depth of a Deep Neural Network and $N_h$ the
number of neurons in the $h$-th hidden layer. We denote by $X^0 \in \mathbb{R}^{N_0}$ the
inputs of the network, $W^h \in \mathbb{R}^{{N_{h}\times{N_{h-1}}}}$ the weights matrix
defining the synaptic strengths between the hidden layer $h$ and its
preceding $h-1$. The output $Y^h$ of the hidden layer $h$ are thus defined as follows:
 
\begin{equation}\label{fp}
X^0   =   x,\
Y^h  =   W^hX^{h-1},\
X^h   =  \sigma(Y^h),\
h   =   1,2,\cdots, L.
\end{equation}
Where $\sigma(.)$ is
a point-wise differentiable activation function. We will thus
denote by $\sigma'(.)$  its first order derivative,  $x\in \mathbb{R}^{N_0}$ is the input to the network and $W^h$ is the weight matrix
of the $h$-th layer. To optimize the neural network, we compute the partial derivatives of the loss $J(f (x), y)$ w.r.t. the weight matrices $\frac{\partial J(f (x),y)}{\partial W^h}$, with $f(x)$ and $y$ are the output of the DNN and the associated target/label respectively. This quantity can be computed similarly by  the following two-step rule:
\begin{equation}\label{bp} 
X^{L}_*   =  \displaystyle\frac{\partial J(f (x),y)}{\partial X^L},\ 
Y^{h}_*  =   \displaystyle X^{h}_*\odot \sigma'(Y^h),\  
X^{h-1}_*   =  \displaystyle\left(W^h\right)^\top Y^{h}_*,\ 
h   =    L,\cdots, 1.
 \end{equation}
Once $Y^{h}_*$ is computed, the weights update can be computed as

\begin{equation}\label{grad1}
\frac{\partial J(f (x),y)}{\partial W^h}=Y^{h}_*\left(X^{h-1} \right)^\top.
\end{equation} 
\end{theorem} 
Hereafter, the two-step recursive formulae given in  \eqref{bp} and \eqref{grad1} will be referred  to as  "the two-step rule for back-propagation" (see the mapping diagram  represented by  Figure \ref{fig:two-step rule}.). In particular, this rule   provide us a simplified formulation of  the so-called "generalized delta rule" similarly to \eqref{feedforward-formulas-scalar1}. Thus, we have answered   our key question.

 \vspace*{2.5cm}
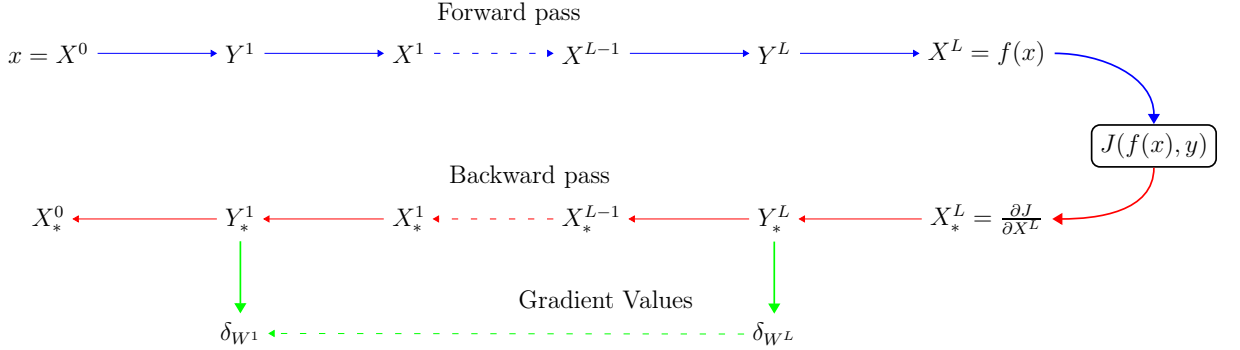
\begin{figure}[htp] 
    	\centering
\begin{tikzpicture}[thick,   transform canvas={scale=0.80}]

\matrix[row sep=2cm,column sep=2.cm] (m1) {
    \node [align=center, xshift=1.5cm] (node21) {$x=X^{0}$}; &
    \node [align=center, xshift=1.5cm] (node22) {$Y^{1}$}; &
    \node [align=center, xshift=1.5cm] (node23) {$X^{1}$}; &
    \node [align=center, xshift=1.5cm] (node24) {$X^{L-1}$};&
    \node [align=center, xshift=1.5cm] (node25) {$Y^{L}$}; &
    \node [align=center, xshift=1.5cm] (node26) {$X^{L}=f(x)$}; 
    \\
    \node [align=center, xshift=1.5cm] (node31) {$X^{0}_*$}; &
    \node [align=center, xshift=1.5cm] (node32) {$Y^{1}_*$}; &
    \node [align=center, xshift=1.5cm] (node33) {$X^{1}_*$}; &
     \node [align=center, xshift=1.5cm] (node34) {$X^{L-1}_*$}; &
    \node [align=center, xshift=1.5cm] (node35) {$Y^{L}_*$};&
    \node [align=center, xshift=1.5cm] (node36) {$X^{L}_*=\frac{\partial J}{\partial X^L}$}; 
    \\
   };
\draw [-Triangle, line width=0.01mm,  blue] (node21) -- (node22);
\draw [-Triangle, line width=0.01mm,  blue] (node22) -- (node23);
\draw [-Triangle, line width=0.01mm,  blue, loosely dashed] (node23) -- (node24);
\draw [-Triangle, line width=0.01mm,  blue] (node24) -- (node25);
\draw [-Triangle, line width=0.01mm,  blue] (node25) -- (node26);
\draw [-Triangle, line width=0.01mm,  red] ([xshift=.01cm]node36.west) -- ([xshift=-.01cm]node35.east);
\draw [-Triangle, line width=0.01mm,  red] ([xshift=.01cm]node35.west) -- ([xshift=-.01cm]node34.east);
\draw [-Triangle, line width=0.01mm,  red, loosely dashed] ([xshift=.01cm]node34.west) -- ([xshift=-.01cm]node33.east);
\draw [-Triangle, line width=0.01mm,  red] ([xshift=.01cm]node33.west) -- ([xshift=-.01cm]node32.east);
\draw [-Triangle, line width=0.01mm,  red] ([xshift=.01cm]node32.west) -- ([xshift=-.01cm]node31.east);
\node[above right = 0mm and -1mm of node23, align=center] (textforward) {Forward pass};
\node[above right = 0mm and 1mm of node33, align=center] {Backward pass};
\node[rounded corners, draw, below right = .8cm and 0.6cm of node26] (nodec) {  $J(f(x),y)$};

\path (node26.east)edge [-Triangle,out=0,in=90, blue] node [midway, anchor=center] {} (nodec.north);

\path (nodec.south)edge [-Triangle,out=270,in=0, red] node [midway, anchor=center] {} (node36.east);
\node[below right = .5cm and -9.2cm of m1, align=center] {Gradient Values};
\node [below =1.2cm of node32, node distance = 2.9cm] (node10_31) {$\delta_{W^{1}}$};
\draw [-Triangle, green] (node32) -- (node10_31);
\node [below =1.2cm of node35, node distance = 2.9cm] (node10_35) {$\delta_{W^{L}}$};
\draw [-Triangle, green] (node35) -- (node10_35);
\draw [-Triangle, line width=0.01mm,  green, loosely dashed] ([xshift=.01cm]node10_35.west) -- ([xshift=-.01cm]node10_31.east);
\end{tikzpicture}
\vspace*{3.5cm}
\caption[ ]{Mapping diagram associated to the two-step rule for backpropagation.}
	\label{fig:two-step rule}
\end{figure}
\begin{proof}
~~~Firstly, for any $h=1,\cdots, L$ let us recall the   simplified notations introduced by \eqref{notations}:

 $$X^{h}_*=\frac{\partial J(f (x),y)}{\partial {X^h}},\ Y^{h}_*=\frac{\partial J(f (x),y)}{\partial {Y^h}}.$$ Secondly, for fixed $h\in\{1,\cdots, L\}$,
  $X^h=\sigma(Y^h)$, then   \eqref{chain:2} implies that
 \begin{equation*}\label{p1}
 \frac{\partial J(f (x),y)}{\partial Y^h}=\frac{\partial J(f (x),y)}{\partial X^h}\odot\sigma'(Y^h)
\end{equation*}
thus
\begin{equation}\label{p2}
 Y^{h}_*=X^{h}_*\odot\sigma'(Y^h).
\end{equation}
On the other hand, $Y^h=W^hX^{h-1}$, thus 
\begin{equation*}\label{p41}
\frac{\partial J(f (x),y)}{\partial X^{h-1}}=(W^h)^\top\frac{\partial J(f (x),y)}{\partial Y^h} 
\end{equation*}
by vertue of \eqref{chain:}. As consequence,

\begin{equation}\label{p5}
X^{h-1}_*=(W^h)^\top Y^{h}_*. 
\end{equation}
Equations \eqref{p2} and \eqref{p5} implies  immediately  \eqref{bp}. Moreover, we apply again  \eqref{chain:} to the cost function $J$ and the relation 
$Y^h=W^hX^{h-1}$, we deduce that
\begin{equation*}\label{p6}
\frac{\partial J(f (x),y)}{\partial W^{h}}=\frac{\partial J(f (x),y)}{\partial Y^h}(X^{h-1})^\top=Y^{h}_*(X^{h-1})^\top.
\end{equation*}
 This end the proof of the Theorem \ref{thm1}. Furthermore, in the practical setting mentioned in  Remark \ref{rem1}, one should  replace ${W^{h}}$ by ${W_\sharp^{h}}$ by vertue of \eqref{usefl-w-sharpe}. Thus,  we have for all 
$
h \in\{L,\cdots, 2\}$

\begin{equation*}
\label{tsr2}
X^{h-1}_* =  \displaystyle({W_\sharp^{h}})^\top Y^{h}_*,
\end{equation*}
 and then the associated two-step rule is  given by
 \begin{equation}\label{bp2} 
X^{L}_*   =  \displaystyle\frac{\partial J(f (x),y)}{\partial X^L},\ 
Y^{h}_*  =   \displaystyle X^{h}_*\odot \sigma'(Y^h),\  
X^{h-1}_*   =  \displaystyle\left(W_\sharp^h\right)^\top Y^{h}_*,\ 
h   =    L,\cdots, 1.
 \end{equation}

\end{proof}
\section{The F-propagation and F-adjoint}\label{B}
~~~ Based on the above idea of  two-scale rule for back-prpagation, we shall introduce  the following two natural definitions :

\begin{definition}[An F-propagation]\label{def1}
{}\noindent

Le $X^0\in\mathbb{R}^{N_0}$ be a given data, $\sigma$ be a coordinate-wise map from $\mathbb{R}^{N_h}$ into $\mathbb{R}^{N_{h+1}}$ and $W^h\in \mathbb{R}^{{N_{h}}\times{N_{h-1}}}$ for all ${1\leq h\leq L}$. We say that we have a two-step recursive F-propagation   $F$  through the DNN $A[N_0,\cdots, N_L]$ if   one has the following family of vectors
\begin{equation}
\label{F-def}
F:=\begin{Bmatrix}X^{0},
Y^{1},X^{1},\cdots,X^{L-1},  Y^{L},  X^{L}
\end{Bmatrix}
\end{equation} with
\begin{equation}
\label{F-def-eq1}
Y^h=W^hX^{h-1}, \ X^h=\sigma(Y^h),\ X^h\in\mathbb{R}^{N_h},\ h=1,\cdots, L.
\end{equation}
\end{definition}
Before going further, let us point that in the above definition the prefix "F" stands for "Feed-forward".
\begin{definition}[The F-adjoint of an F-propagation]\label{def2}
{}\noindent

Let $X^{L}_{*}\in\mathbb{R}^{N_L}$ be a given data and  a two-step recursive F-propagation  $F$ through the DNN $A[N_0,\cdots, N_L]$. We define the F-adjoint     ${F}_{*}$ of the F-propagation  $F$  as follows
\begin{equation}
\label{F*-def}
F_{*}:=\begin{Bmatrix}X^{L}_{*},
Y^{L}_{*}, X^{L-1}_{*},\cdots, X^{1}_{*},Y^{1}_{*}, X^{0}_{*}
\end{Bmatrix}
\end{equation} with
\begin{equation}
\label{F*-def-eq1}
Y^h_{*}=X^{h}_{*}\odot {\sigma}'(Y^h), \ X^{h-1}_{*}=(W^h)^\top Y^h_{*},\ h=L,\cdots, 1.
\end{equation}
\end{definition}
 \begin{remark}
 \noindent
 
 It is immediately seen that if for every $h=1,\cdots,L$, $W^h$ is  an orthogonal matrix i.e. $({W^h})^\top W^h=W^h({W^h})^\top =I_{N_h}$, $\sigma(X)=X$ and $X^0=X^0_*$ one  has 
$$Y^{1}={W^{1}}X^{0}_*={W^{1}}({W^{1}})^\top Y^{1}_*= Y^{1}_*,$$
  $$Y^{2}={W^{2}}X^{1}_*={W^{2}}({W^{2}})^\top Y^{2}_*= Y^{2}_*,$$
  by recurrence we obtain
  $$Y^{L}={W^{L}}X^{L-1}_*={W^{L}}({W^{L}})^\top Y^{L}_*= Y^{L}_*.$$  
On the other hand, 
  $$ X^1=\sigma({W^L}X^{0})={W^L}X^{0}_*= Y^{1}_*= X^{1}_*,$$ 
  also, by recurrence we get
  
  $$ X^{L}={W^{L}} X^{L-1}={W^L}({W^{L}})^\top Y^{L}_*=Y^L_*=X^L_*.$$
 Consequently, we obtain $F_*=F$ and we  refer to this property as F-symmetric. We may conjecture   that under the following assumption : for all vector $X$, $\sigma(X)=X$, a DNN is F-symmetric if and only if $({W^h})^\top W^h=W^h({W^h})^\top =I_{N_h}, h=1,\cdots L$, i.e. for all $h=1,\cdots ,L$, $W^h$ is an orthogonal matrix.
 \end{remark}

Similarly to the note at the end of the proof of Theorem \ref{thm1}, if we are  in the practical setting mentioned in  Remark \ref{rem1}, one should  replace ${W^{h}}$ by ${W_\sharp^{h}}$ by vertue of \eqref{usefl-w-sharpe} and \eqref{F*-def-eq1} by the following relations :

 \begin{equation}\label{bp22} 
Y^{h}_*  =   \displaystyle X^{h}_*\odot \sigma'(Y^h),\  
X^{h-1}_*   =  \displaystyle\left(W_\sharp^h\right)^\top Y^{h}_*,\ 
h   =    L,\cdots, 1.
 \end{equation}
 
Now, following these definitions, we can deduce our second main mathematical result.

\begin{theorem}[The backpropagation is an F-adjoint of the associated F-propagation]
{}\noindent\label{thm2}

Let  $A[N_0,\cdots, N_L]$ be   a Deep Neural Network and let $F:=\begin{Bmatrix}X^0,
Y^{1},X^{1},\cdots,X^{L-1},  Y^{L},  X^{L}
\end{Bmatrix} $  an F-propagation through $A[N_0,\cdots, N_L]$, with  weights matrix $W^h \in \mathbb{R}^{{N_{h}\times{N_{h-1}}}}$  and fixed point-wise activation $\sigma^h=\sigma$, $h=1,\cdots, L$.  

Let $X^0=x\in\mathbb{R}^{N_0}$   be the input to   this network and  $J(f (x), y)$ be the loss function, with $f(x)$ and $y$ are the output of the DNN and the associated target/label respectively.

Then the associated backpropagation pass is computed with  the F-adjoint of $F$ with $X^{L}_{*}=\frac{\partial J}{\partial X^L}(f (x), y)$, that is 
$  F_{*}:=\begin{Bmatrix}X^{L}_{*},
Y^{L}_{*}, X^{L-1}_{*},\cdots, X^{1}_{*},Y^{1}_{*}, X^{0}_{*}
\end{Bmatrix}$. Moreprecisely, for all $h=1, \cdots, L$
\begin{equation*}\label{p4}
\frac{\partial J(f (x),y)}{\partial W^{h}}=Y^{h}_{*}(X^{h-1})^\top.
\end{equation*}
\end{theorem}
Before proving the Theorem \ref{thm2}, let us as remark that this mathematical result is a simplified version of Theorem  \ref{thm1} based on the notion of F-adjoint introduced in Definition \ref{def2}. 
\begin{proof}

By definition of the F-adjoint one has for all $h=L,\cdots, 1,$  
$$Y^h_{*}=X^{h}_{*}\odot {\sigma}'(Y^h), \ X^{h-1}_{*}=(W^h)^\top Y^h_{*}.$$
Moreover, by the two-step rule for backpropagation, we have 
$$
X^{L}_{*}  =  \displaystyle\frac{\partial J(.)}{\partial X^L},\ 
Y^h_{*}  =  \displaystyle X^h_{*}\odot\sigma'(Y^h),\ 
\delta_{W^{h}}  =  \displaystyle Y^h_{*}(X^{h-1})^\top
,\ 
X^{h-1}_{*} =  \displaystyle({W^{h}})^\top Y^h_{*},
$$
As consequence, we deduce that the backpropagation is exactely determined by the F-adjoint $F_*$ relatively to the choice of $ X^{L}_{*}=\displaystyle\frac{\partial J(.)}{\partial X^L}$. Here we present a new proof of Theorem 1 of \cite{bougham2023} which is simpler than the original one.
\end{proof}
\section{Application to  the two simplest cases   $A[1, 1, 1]$ and $A[1, 2, 1]$}
~~~The present section shows that in the following  four simplest cases associated to the DNN $A[1, N_1, 1]$ with $N_1=1,2$, we shall apply the notion of F-adjoint   to compute the partial derivative of the elementary cost function $J$ defined by $J(f(x),y)=f(x)-y$ for any real $x$ and fixed real $y$. In this particular setting, we have the two simplest cases $A[1, 1, 1]$ and $A[1, 2, 1]$:  one neuron and two neurons in the hidden layer (see Figures \ref{fig:case1} and  \ref{fig:case3}). 
\subsection{The first case: $A[1, 1, 1]$}
~~~   The first simplest case corresponds to  
$A[1, 1, 1]$ architecture is shows by the Figure \ref{fig:case1}. Let us denote by a $W^{1}=\begin{pmatrix}
\alpha_{11}^{1} &\alpha_{12}^{1}
\end{pmatrix}$ and   $W^{2}=\begin{pmatrix}
\alpha_{1}^{2} &\alpha_{2}^{2}
\end{pmatrix}$ the weights in the first and second layer. We will evaluate the $\delta_{W^1}$ and $\delta_{W^2}$ by the differential calculus rules firstly and then recover this result by the two-step rule for back-propagation.
\tikzstyle{inputNode}=[draw,circle,minimum size=30pt,inner sep=0pt]
\tikzstyle{stateTransition}=[-stealth, thick]
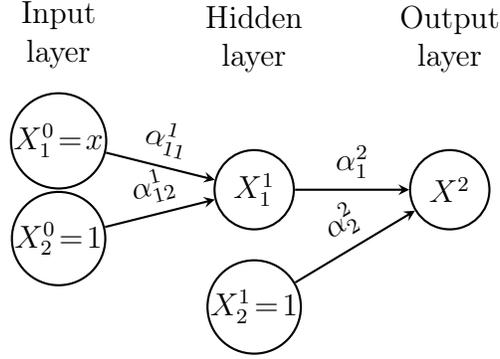
\begin{figure}[htp]
	\centering

\begin{tikzpicture}[scale=1.3]
	\node[inputNode, thick] (i1) at (6, 0.5) {$X_1^0\!=\!x$};	
	\node[inputNode, thick] (i3) at (6, -0.5) {$X_2^0\!=\!1$};	
	\node[inputNode, thick] (h1) at (8, 0) {$X_1^1$};
	\node[inputNode, thick] (h6) at (8, -1.2) {$X_2^1\!=\!1$};	
	\node[inputNode, thick] (o1) at (10, 0) {$X^2$};
	\draw[stateTransition] (i1) -- (h1) node [midway, sloped, above] {$\alpha_{11}^{1}$};	
		\draw[stateTransition] (i3) -- (h1) node [midway, sloped, above] {$\alpha_{12}^{1}$};
	\draw[stateTransition] (h1) -- (o1) node [midway, sloped, above] {$\alpha_{1}^{2}$};
	\draw[stateTransition] (h6) -- (o1) node [midway, sloped, above] {$\alpha_{2}^{2}$};	
	
	\node[above=5pt of i1, align=center] (i1) {Input \\ layer};
	\node[above=25pt of h1, align=center] (h1) {Hidden \\ layer};
	\node[above=25pt of o1, align=center] (o1) {Output \\ layer};
	
	
\end{tikzpicture}

\caption[Case 1.]{The DNN associated to the case $1$.}
	\label{fig:case1}
\end{figure}
\subsubsection*{The F-propagation through this DNN is given by :}

Obviously, we have
$$X^{0}=\begin{pmatrix}x\\
1 \end{pmatrix},\ 
Y^1=\alpha_{11}^{1}x+\alpha_{12}^{1}, \ 
X^1=\begin{pmatrix}\sigma(\alpha_{11}^{1}x+\alpha_{12}^{1})\\
1
\end{pmatrix},\ $$
$$Y^2=\alpha_{1}^{2}\sigma(\alpha_{11}^{1}x+\alpha_{12}^{1})+\alpha_{2}^{2},\ X^2=\sigma\Bigl[\alpha_{1}^{2}\sigma(\alpha_{11}^{1}x+\alpha_{12}^{1})+\alpha_{2}^{2}\Bigr].
$$
Let us denote $y^1:=\alpha_{11}^{1}x+\alpha_{12}^{1}$ and  $y^2:= \alpha_{1}^{2}\sigma(y^1)+\alpha_{2}^{2}$, then 
 $$F=\Biggl\{X^{0}=\begin{pmatrix}x\\
1 \end{pmatrix},\ 
Y^1=y^1, \ 
X^1=\begin{pmatrix}\sigma(y^1)\\
1
\end{pmatrix},\ Y^2=y^2,\ X^2=\sigma(y^2)
\Biggr\}$$ and $$J(f(x),y)=X^2-y=\sigma(y^2)-y=\sigma\Bigl[\alpha_{1}^{2}\sigma(\alpha_{11}^{1}x+\alpha_{12}^{1})+\alpha_{2}^{2}\Bigr]-y.$$
Hence, by using the differential calculus rules one gets

\begin{equation}
\label{ex11}
 \delta_{W^2}=\begin{pmatrix}\frac{\partial J}{\partial {\alpha_{1}^{2}}}\\[8pt]
\frac{\partial J}{\partial {\alpha_{2}^{2}}} \end{pmatrix}^\top
 =\begin{pmatrix}{\sigma'(y^2)}\sigma(y^1)\\[8pt]
 {\sigma'(y^2)}
\end{pmatrix}^\top,\   
\delta_{W^1}=\begin{pmatrix}\frac{\partial J}{\partial {\alpha_{11}^{2}}}\\[8pt]
\frac{\partial J}{\partial {\alpha_{12}^{2}}} \end{pmatrix}^\top=\begin{pmatrix}
{\sigma'(y^2)}{\sigma'(y^1)}\alpha_{1}^{2}x\\[8pt]
{\sigma'(y^2)}\sigma'(y^1)\alpha_{1}^{2}
\end{pmatrix}^\top.\end{equation}
\subsubsection*{The F-adjoint of the above F-propagation  is given   by}
$$F_*=\Biggl\{X^{2}_{*}=1,\ 
Y^{2}_{*}={\sigma'(y^2)},\ 
X^{1}_{*}=\alpha_1^2{\sigma'(y^2)},\ 
Y^{1}_{*}=\alpha_1^2{\sigma'(y^2)}{\sigma'(y^1)},\
X^{0}_{*}=\alpha_{11}^1\alpha_1^2{\sigma'(y^2)}{\sigma'(y^1)}
\Biggr\}$$
since by \eqref{bp22} we have 
$$ 
Y^{h}_*  =   \displaystyle X^{h}_*\odot \sigma'(Y^h),\  
X^{h-1}_*   =  \displaystyle\left(W_\sharp^h\right)^\top Y^{h}_*,\ 
h   =    2, 1 
$$
then we deduce that
$$ Y^{2}_*  =X^{2}_*\odot \sigma'(y^2)= \sigma'(y^2),\  X^{1}_*   =   \left(W_\sharp^2\right)^\top Y^{2}_*=\alpha_1^2 \sigma'(y^2), \ 
Y^{1}_*  =X^{1}_*\odot \sigma'(y^1)=\alpha_1^2 \sigma'(y^2)\sigma'(y^1)$$
and
$$X^{0}_*   =   \left(W_\sharp^1\right)^\top Y^{1}_*=\alpha_{11}^1Y^{1}_*=
\alpha_{11}^1\alpha_1^2 \sigma'(y^2)\sigma'(y^1)$$


Hence, by using the F-adjoint and the two-step rule \eqref{bp2} one gets

$$
 \delta_{W^2}=Y^{2}_{*}(X^1)^\top =\begin{pmatrix}{\sigma'(y^2)}\sigma(y^1)\\[8pt]
 {\sigma'(y^2)}
\end{pmatrix}^\top,  \ 
\delta_{W^1}=Y^{1}_{*}(X^0)^\top =\begin{pmatrix}
{\sigma'(y^2)}{\sigma'(y^1)}\alpha_{1}^{2}x\\[8pt]
{\sigma'(y^2)}\sigma'(y^1)\alpha_{1}^{2}
\end{pmatrix}^\top.$$ Thus we recover, via the F-adjoint $F_*$, the computation given by \eqref{ex11}.
\subsection{The second case: $A[1, 2, 1]$}

~~~   The second simplest  case corresponds to  
$A[1, 2, 1]$ architecture is shows by the Figure \ref{fig:case3}.
In this case we have
$W^{1}=\begin{pmatrix}
\alpha_{11}^{1} &\alpha_{12}^{1}\\[5pt]
\alpha_{21}^{1}& \alpha_{22}^{1}
\end{pmatrix}$ and   $W^{2}=\begin{pmatrix}
\alpha_{1}^{2} &\alpha_{2}^{2}& \alpha_{3}^{2}
\end{pmatrix}$. Thus, one deduce immediately that 
 
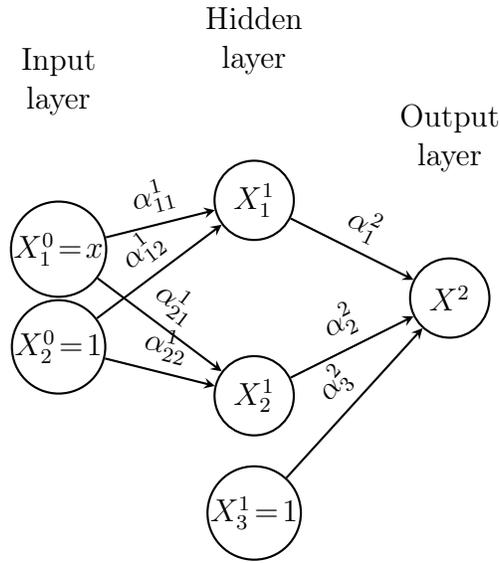
\begin{figure}[htp]
	\centering

\begin{tikzpicture}[scale=1.3]
	\node[inputNode, thick] (i1) at (6, 0.5) {$X_1^0\!=\!x$};
	
	\node[inputNode, thick] (i3) at (6, -0.5) {$X_2^0\!=\!1$};
	
	\node[inputNode, thick] (h1) at (8, 1) {$X_1^1$};
	
	\node[inputNode, thick] (h5) at (8, -1) {$X_2^1$};
	\node[inputNode, thick] (h6) at (8, -2.2) {$X_3^1\!=\!1$};
	
	\node[inputNode, thick] (o1) at (10, 0) {$X^2$};

	
	
	\draw[stateTransition] (i1) -- (h1) node [midway, sloped, above] {$\alpha_{11}^{1}$};
	
	\draw[stateTransition] (i1) -- (h5) node [midway, sloped, above] {$\alpha_{21}^{1}$};
		\draw[stateTransition] (i3) -- (h1) node [midway, sloped, above] {$\alpha_{12}^{1}$};
		\draw[stateTransition] (i3) -- (h5) node [midway, sloped, above] {$\alpha_{22}^{1}$};
	
	\draw[stateTransition] (h1) -- (o1) node [midway, sloped, above] {$\alpha_{1}^{2}$};
		\draw[stateTransition] (h5) -- (o1) node [midway, sloped, above] {$\alpha_{2}^{2}$};
	\draw[stateTransition] (h6) -- (o1) node [midway, sloped, above] {$\alpha_{3}^{2}$};

	\node[above=of i1, align=center] (l1) {Input \\ layer};
	\node[above=of h1, align=center] (h1) {Hidden \\ layer};
	\node[above=of o1, align=center] (o1) {Output \\ layer};
	
	
\end{tikzpicture}

\caption[Case 3.]{The  DNN associated to the case $2$.}
	\label{fig:case3}
\end{figure}
\subsubsection*{The F-propagation through this DNN is given by :}
$$F=\Biggl\{
X^{0}=\begin{pmatrix}x\\
1 \end{pmatrix}
,\ 
Y^1= \begin{pmatrix}y_{11}^1\\ 
y_{21}^1\end{pmatrix} 
,\ 
X^1 =\begin{pmatrix}\sigma(y_{11}^1)\\ 
\sigma(y_{21}^1)\\ 
1
\end{pmatrix}
,\
Y^2 =y^2
,\ 
X^2 =
\sigma(y^2)
\Biggr\}$$
Hence, by using the differential calculus rules one gets

$$
 \delta_{W^2}=\begin{pmatrix} \sigma'(y^2)\sigma(y_{11}^1)\\[7pt]
 {\sigma'(y^2)}\sigma(y_{21}^1)\\[7pt]
 {\sigma'(y^2)}
\end{pmatrix}^\top,\ 
\delta_{W^1}=\begin{pmatrix}\alpha_1^2x{\sigma'(y_{11}^1)}{\sigma'(y_{21}^1)} & \alpha_1^2{\sigma'(y_{11}^1)}{\sigma'(y^2)}\\[7pt]
\alpha_2^2x{\sigma'(y_{21}^1)}{\sigma'(y^2)} & \alpha_2^2{\sigma'(y_{21}^1)}{\sigma'(y^2)}
\end{pmatrix}$$ with

$$y^2:=\alpha_{1}^{2}\sigma(w^1x)+\alpha_{2}^{2}\sigma(w^2)+\alpha_{3}^{2},\  
  y_{11}^1:=\alpha_{11}^{1}x+\alpha_{12}^{1},\  y_{21}^1:=\alpha_{21}^{1}x+\alpha_{22}^{1}.$$
\subsubsection*{The F-adjoint of the above F-propagation  is given   by}

$$F_*=\Biggl\{
X^{2}_{*}=1,\ Y^{2}_{*}={\sigma'(y^2)},\  
X^{1}_{*}=\begin{pmatrix} \alpha_1^2\sigma'(y^2)\\[8pt]
\alpha_2^2\sigma'(y^2)
\!\end{pmatrix},\ 
Y^{1}_{*}=\begin{pmatrix} \alpha_1^2\sigma'(y^2)\sigma'(y_{11}^1)\\[8pt]
\alpha_2^2\sigma'(y^2)\sigma'(y_{21}^1)
\end{pmatrix},\ X^{0}_{*}=x^0\Biggr\}$$ with $x^0=\alpha_{11}^1\alpha_1^2\sigma'(y^2)\sigma'(y_{11}^1)
+\alpha_{21}^1\alpha_2^2\sigma'(y^2)\sigma'(y_{21}^1)$. As above, by using also the F-adjoint and the two-step rule \eqref{bp2} one gets

$$
 \delta_{W^2}=Y^2_*(X^1)^\top=\begin{pmatrix} {\sigma'(y^2)}\sigma(y_{11}^1)\\[8pt]
 {\sigma'(y^2)}\sigma(y_{21}^1)\\[8pt]
 {\sigma'(y^2)}
\end{pmatrix}^\top,\ 
\delta_{W^1}=Y^1_*(X^0)^\top=\begin{pmatrix}\alpha_1^2x{\sigma'(y_{11}^1)}{\sigma'(y^2)} & \alpha_1^2{\sigma'(y_{11}^1)}{\sigma'(y_{21}^1)}\\[8pt]
\alpha_2^2x{\sigma'(y_{21}^1)}{\sigma'(y_{21}^1)} & \alpha_2^2{\sigma'(y_{21}^1)}{\sigma'(y_{21}^1)}
\end{pmatrix}.$$
\section{Related works and Conclusion}
~~~ To the best of our knowledge, in literature, the related  works to this paper  are \cite{Alber2018} and \cite{Hojabr2020}. In particular, in the first paper the authors uses some decomposition of the partial derivatives of the cost function, similar to the two-step formula (cf. \eqref{bp}), to replace the  standard  back-propagation. In addition, they show (experimentally) that for specific scenarios, the two-step decomposition yield better generalization performance than the   one based on the  standard  back-propagation. But in the second article, the authors find some similar update equation similar to the one given by \eqref{bp} that report similarly to standard back-propagation at convergence. Moreover, this method discovers new variations of the back-propagation  by learning new propagation rules that optimize the generalization performance after a few epochs of training.  

In conclusion, we have provided a two-step rule for back-propagation similar to the one for forward propagation and a new   mathematical notion called F-adjoint which is combined by the  two-step rule for back-propagation   describes, in a simple and direct way,  the computational process given by the  back-propagation pass. We hope that, the power and simplicity of the F-adjoint concept  may   inspire the exploration of novel   approaches for optimizing some artificial neural networks training algorithms and investigating some mathematical properties of the F-propagation and F-adjoint notions.  As future work, we envision to explore some   mathematical results regarding  the   F-adjoint for deep neural networks  with respect to the choice of the activation function.

    

\section*{Conflicts of Interest}
The author declare no conflict of interest.

\bibliographystyle{ieeetr}
\bibliography{bougham-sn-article}

\begin{thebibliography}{1}

\bibitem{McCulloch1943}
W.~S. McCulloch and W.~Pitts, ``A logical calculus of the ideas immanent in
  nervous activity,'' {\em The bulletin of mathematical biophysics}, vol.~5,
  pp.~115--133, 1943.

\bibitem{Baldi2021}
P.~Baldi, {\em Deep learning in science}.
\newblock Cambridge: Cambridge University Press, 2021.

\bibitem{Schmidhuber15}
J.~Schmidhuber, ``Deep learning in neural networks: An overview,'' {\em Neural
  networks}, vol.~61, pp.~85--117, 2015.

\bibitem{rumelhart1986}
D.~E. Rumelhart, G.~E. Hinton, and R.~J. Williams, ``Learning representations
  by back-propagating errors,'' {\em nature}, vol.~323, no.~6088, pp.~533--536,
  1986.

\bibitem{bougham2023}
A.~Boughammoura, ``A two-step rule for backpropagation,'' 2023.
\newblock Preprint at \url{https://arxiv.org/abs/2304.13537}.

\bibitem{Ye2022}
J.~C. Ye, {\em Geometry of Deep Learning}.
\newblock Heidelberg: Springer, 2022.

\bibitem{Alber2018}
M.~Alber, I.~Bello, B.~Zoph, P.-J. Kindermans, P.~Ramachandran, and Q.~Le,
  ``Backprop evolution,'' 2018.
\newblock Preprint at \url{https://arxiv.org/abs/1808.02822}.

\bibitem{Hojabr2020}
R.~Hojabr, K.~Givaki, K.~Pourahmadi, P.~Nooralinejad, A.~Khonsari, D.~Rahmati,
  and M.~H. Najafi, ``Taxonn: A light-weight accelerator for deep neural
  network training,'' in {\em 2020 IEEE International Symposium on Circuits and
  Systems (ISCAS)}, pp.~1--5, IEEE, 2020.

\end{thebibliography}

\end{document}